\documentclass[10pt,twocolumn,letterpaper]{article}

\newcommand{\x}{\mathbf{x}}
\newcommand{\s}{\mathbf{s}}
\newcommand{\1}{\mathbf{1}}
\newcommand{\w}{\mathbf{w}}
\newcommand{\bt}{\mathbf{t}}
\newcommand{\W}{\mathbf{W}}
\newcommand{\T}{\mathbf{T}}

\newcommand{\Acal}{\mathcal{A}}

\newcommand{\Rcal}{\mathcal{R}}
\newcommand{\Dcal}{\mathcal{D}}
\newcommand{\Vcal}{\mathcal{V}}

\usepackage{cvpr}
\usepackage{times}
\usepackage{epsfig}
\usepackage{graphicx}
\usepackage{amsmath}
\usepackage{amssymb}

\DeclareGraphicsExtensions{.pdf,.jpeg,.png}

\usepackage{booktabs}
\usepackage{epstopdf}
\usepackage{multirow}
\usepackage{cite}
\usepackage{array}
\usepackage{rotating}
\usepackage[caption=false,font=normalsize,labelfont=sf,textfont=sf]{subfig}

\usepackage[breaklinks=true,colorlinks,bookmarks=false]{hyperref}
\usepackage{url}
\cvprfinalcopy % *** Uncomment this line for the final submission

\def\cvprPaperID{2650} % *** Enter the CVPR Paper ID here

\ifcvprfinal\fi
\renewcommand\footnotemark{}

\begin{document}

%%%%%%%%% TITLE
\title{Semantically Consistent Regularization for Zero-Shot Recognition}

\author{Pedro Morgado\thanks{This work was funded by graduate fellowship SFRH/BD/109135/2015 from the Portuguese Ministry of Sciences and Education and NRI Grants IIS-1208522 and IIS-1637941 from the National Science Foundation.} \qquad Nuno Vasconcelos\\
	Department of Electrical and Computer Engineering\\
	University of California, San Diego\\
	{\tt\small \{pmaravil,nuno\}@ucsd.edu}
%\and
%Nuno Vasconcelos\\
}

\maketitle
\thispagestyle{empty}

%%%%%%%%% ABSTRACT
\begin{abstract}
The role of semantics in zero-shot learning is considered. The effectiveness of previous approaches is analyzed according to the form of supervision provided. While some learn semantics independently, others only supervise the semantic subspace explained by training classes. Thus, the former is able to constrain the whole space but lacks the ability to model semantic correlations. The latter addresses this issue but leaves part of the semantic space unsupervised. This complementarity is exploited in a new convolutional neural network (CNN) framework, which proposes the use of semantics as constraints for recognition.Although a CNN trained for classification has no transfer ability, this can be encouraged by learning an hidden semantic layer together with a semantic code for classification. Two forms of semantic constraints are then introduced. The first is a loss-based regularizer that introduces a generalization constraint on each semantic predictor. The second is a codeword regularizer that favors semantic-to-class mappings consistent with prior semantic knowledge while allowing these to be learned from data. Significant improvements over the state-of-the-art are achieved on several datasets.
\end{abstract}

%%%%%%%%% BODY TEXT
\section{Introduction}
\label{sec:intro}

Significant advances in object recognition have been recently achieved with the introduction of deep convolutional neural networks (CNNs). The main limitation of this approach is the effort required to 1) collect and annotate millions of images necessary to train these models, and 2) the complexity of training a CNN from scratch. In fact, most recent computer vision papers use or adapt a small set of popular models, such as AlexNet~\cite{Krizhevsky2012}, GoogLeNet~\cite{Szegedy2014}, and VGG~\cite{Simonyan2014}, learned from the Imagenet dataset~\cite{Deng2009}. Hence, there is an interest in techniques for transfer learning, where a model learned on a dataset is used to recognize object classes that are not represented in it. Ideally, transfer learning methods would replicate the human ability to recognize objects from a few example images or even from a description in terms of concepts in some semantic vocabulary.

This has motivated the introduction of semantic representations for object recognition \cite{Rasiwasia2007,Vogel2007,Li2010,Rasiwasia2012,Torresani2010}, which rely on a predefined vocabulary of visual concepts to define a semantic space $\cal S$ and a set of classifiers to map each image into that space. The scores of these classifiers can then be used as semantic features for object classification. Furthermore, because simple rules of thumb can be designed, a priori, to describe new object classes in terms of these semantics, the image mapping into $\cal S$ can be exploited to recognize previously unseen objects. This is known as {\it zero-shot learning\/} (ZSL) \cite{Lampert2009,Akata2015,Romera2015,Farhadi2009,Rohrbach2011,Akata2013}. 

The fundamental difficulty of ZSL is that training cannot be guided by the end goal of the classifier. While the recognizer is learned from a set of {\it training\/} classes, it must provide accurate predictions for image classification into a non-overlapping set of {\it unseen\/} or {\it zero-shot\/} (ZS) classes. Historically, early efforts were devoted to the identification of good semantics for ZSL. This motivated the collection of datasets containing images annotated with respect to semantics such as {\it visual  attributes\/} \cite{Lampert2009,Farhadi2009}. Subsequent works addressed the design of the semantic space $\cal S$, using one of two strategies previously proposed in the semantic representation literature. The first, {\it recognition using independent semantics\/} (RIS), consists of learning an independent classifier per semantic \cite{Vogel2007,Li2010,Torresani2010}. Due to its simplicity, RIS became widely popular in the attribute recognition literature \cite{Lampert2009,	Farhadi2009,Su2010,Rohrbach2011,Parikh2011,Wang2013}. Notwithstanding efforts in discriminant attribute discovery \cite{Chen2014,Kumar2009,Farhadi2009,Parikh2011,Rastegari2012} or modeling of uncertainty \cite{Jayaraman2014,Lampert2009,Wang2013}, learning semantics independently proved too weak to guarantee reliable ZS predictions.

This motivated a shift to the second strategy, which ties the design of $\cal S$ to the goal of recognition, by learning a single multi-class classifier that optimally discriminates between all training classes \cite{Rasiwasia2007,Rasiwasia2012}. 
The difficulty of extending this approach to ZSL is that the semantics of interest are not the classes themselves. 
\cite{Akata2013} proposed an effective solution to this problem by noting that there is a fixed linear transformation, or embedding, between the semantics of interest and the class labels, which can be specified by hand, even for ZS classes.
This was accomplished using a label embedding function $\phi$, to map each class $y$ into a vector $\phi(y)$ in the space of attributes.
Recently, various works have proposed variations on this approach~\cite{Akata2015,Li2015,Romera2015,Qiao2016,Reed2016,Akata2016b}. We refer to this class of methods as {\it recognition using semantic embeddings\/} (RULE). By learning all semantics simultaneously, RULE is able to leverage dependencies between concepts, thus addressing the main limitation of RIS.

In this work, we investigate the advantages and disadvantages of the two approaches for implementations based on deep learning and CNNs. We show that, in this context, the two methods reduce to a set of {\it constraints\/} on the CNN architecture: RIS learns a bank of independent CNNs, and RULE uses a single CNN with fixed weights in the final layer. It follows that the performance of the two approaches is constrained by the form in which supervision is provided on the space $\Acal$ of image attributes. While RIS provides supervision along each dimension independently, RULE does so along the subspace  spanned by the label embedding directions $\phi(y)$. Because the number of attributes is usually larger than classes, this exposes the strengths and weaknesses of the two approaches. On one hand, RIS supervises all attributes but cannot model their dependencies. On the other, RULE models dependencies but leaves a large number of dimensions of $\Acal$ unconstrained. 

To exploit this complementarity, we propose a new framework denoted \textit{Semantically COnsistent REgularization} (SCoRe) that leverages the advantages of \textit{both} RIS and RULE. This is achieved by recognizing that the two methods exploit \textit{semantics} as \textit{constraints for recognition}. While RIS enforces first-order constraints (single semantics), RULE focuses second-order (linear combinations). However, both are suboptimal for ZSL. 
%none of the two approaches does this in a way that guarantees best ZSL generalization. 
RIS ignores the recognition of training classes, sacrificing the modeling of semantic dependencies, and RULE ignores a large subspace of $\Acal$ and fixes network weights.
%, which is sub-optimal for learning. 
SCoRe addresses these problems by exploiting the view of a CNN as an optimal classifier with respect to a multidimensional classification code, implemented at the top CNN layer. It interprets this code as a mapping between semantics (layer before last) and classes (last layer). It then enforces {\it both\/} first and second-order regularization constraints through a combination of 1) an RIS like {\it loss-based regularizer\/} that constraints semantic predictions, and 2) a {\it codeword regularizer\/} that favors classification codes consistent with RULE embeddings. %Experimental results shows that this enables significant gains in ZSL performance over the state-of-the-art in various datasets.

\section{Previous Work}
\label{sec:review}

\vspace{-5pt}
\paragraph{Semantics}
Semantics are visual descriptions that convey meaning about an image $\x\in\mathcal{X}$, and may include any measurable visual property: discrete or continuous, numerical or categorical. Given a semantic vocabulary $\mathcal{V} = \{v_1, \ldots, v_Q\}$, a semantic feature space $\mathcal{S}$ is defined as the Cartesian product of the vector spaces $\mathcal{S}_k$ associated with each semantic $v_k$, $\mathcal{S} = \mathcal{S}_1 \times \cdots \times \mathcal{S}_Q$. A classifier is denoted semantic if it operates on $\cal S$. As an example, for animal recognition, a semantic vocabulary containing visual attributes, e.g.~$\Vcal \in \{\textit{furry}, \textit{has legs}, \textit{is brown}, etc.\}$, is usually defined along with their corresponding vector spaces. In this case, since all semantics are binary, ${\cal S}_k=\mathbb{R}$ where large positive values indicate the attribute presence, and large negative values, its absence.

Early approaches to semantic recognition~\cite{Rasiwasia2012} used the set of image classes to be recognized as the semantic vocabulary. The rationale is to create a feature space with a high-level abstraction, where operations such as image search \cite{Rasiwasia2007} or classification \cite{Rasiwasia2012, Li2010} can be performed more robustly. 
More recently, there has been substantial interest in semantic feature spaces for transfer learning, which use an auxiliary semantic vocabulary, defined by mid-level visual concepts. Three main categories of concepts have been explored, including visual attributes, hierarchies and word vector representations. Attributes were introduced in \cite{Lampert2009, Farhadi2009} and quickly adopted in many other works \cite{Rohrbach2011, Su2010, Akata2013, Jayaraman2014, Wang2013, Huang2015, hwang2011, Kodirov2015, Gan2016, Changpinyo2016, Xian2016}. Semantic concepts extracted from hierarchies/taxonomies were later explored in \cite{Akata2013,Akata2015,Rohrbach2011,Xian2016}, and vector representations for words/entities in \cite{Akata2015, Fu2015, Fu2015b, Frome2013, Norouzi2013, Reed2016, Changpinyo2016, Qiao2016, Xian2016}. 

\vspace{-10pt}
\paragraph{Zero-shot learning}
Most current solutions to ZSL fall under two main categories: RIS and RULE. Early approaches adopted the RIS strategy. One of the most popular among these is the direct attribute prediction (DAP) method~\cite{Lampert2009}, which learns attributes \textit{independently} using SVMs and infers ZS predictions by a maximum a posteriori rule that assumes attribute independence. Several enhancements have been proposed to account for attribute correlations {\it a posteriori,\/} e.g. by using CRFs to model attribute/class correlations \cite{Chen2012}, directed Bayesian networks to merge attribute predictions into class scores \cite{Wang2013}, or random forests learned so as to mitigate the effect of unreliable attributes \cite{Jayaraman2014}. More recently, \cite{Liang2015} proposed a multiplicative framework that enables class-specific attribute classifiers, and \cite{AlHalah2015} learns independent attributes which were previously discovered from Word2Vec representations. 

RULE is an alternative strategy that exploits the one-to-one relationship between semantics and object classes. The central idea is to define an embedding $\phi(\cdot)$ that maps each class $y$ into a $Q$-dimensional vector of attribute states $\phi(y)$ that identifies it. A bilinear compatibility function 
\begin{equation}
h(\x,y;\T) = \phi(y)^T \T^T \theta(\x)
\label{eq:bilin}
\end{equation}
of parameters $\T\in\mathbb{R}^{d\times Q}$ is then defined between the feature vector $\theta(\x)\in\mathbb{R}^d$ of image $\x$ and the encoding of its class $y$. In the first implementation of RULE for ZSL \cite{Akata2013}, $\T$ is learned by a variant of the structured SVM.
Several variants have been proposed, such as the addition of different regularization terms \cite{Romera2015,Qiao2016}, the use of least-squares losses for faster training \cite{Romera2015}, or improved semantic representations of objects learned from multiple text sources \cite{Akata2016b,Reed2016}.

\section{Semantics and deep learning}
\label{sec:compare}
\vspace{-2pt}
We now discuss the CNN implementation of RIS and RULE. For simplicity, we assume attribute semantics. Sections~\ref{sec:semantics} and~\ref{sec:deep-score} extend the treatment to other concepts. For quick consultation, Table \ref{tab:notation} summarizes important notation used in the rest of the paper.

\subsection{Deep-RIS}
\vspace{-1pt}
Under the independence assumption that underlies RIS, the CNN implementation reduces to learning $Q$ independent attribute predictors. Inspired by the success of multi-task learning, it is advantageous to share CNN parameters across attributes, and rely on a common feature extractor $\theta(\x; \Theta)$ of parameters $\Theta$, which can be implemented with one of the popular CNNs in the literature. Thus, each attribute predictor $a_k$ of Deep-RIS takes the form
\begin{equation}
a_k(\x;\bt_k,\Theta) = \sigma\left(\bt^T_k \theta(\x;\Theta)\right)
\end{equation}
where $\sigma(\cdot)$ is the sigmoid function and $\bt_k$ a parameter vector. Given a training set $\Dcal = \lbrace(\x^{(i)}, \s^{(i)})_{i=1}^N\rbrace$, where $\s^{(i)} = (s_1^{(i)}, \ldots, s_Q^{(i)})$ are attribute labels, $\bt_k$ and $\Theta$ are learned by minimizing the risk
\begin{equation}
\Rcal[a_1,\ldots,a_Q,\Dcal] = \sum_i\sum_k L_b(a_k(\x^{(i)}; \bt_k,\Theta), s_k^{(i)})
\label{eq:RTISrisk}
\end{equation}
where $L_b$ is a binary loss function, typically the cross-entropy loss $L_b(v,y) = -y \log(v) - (1-y) \log(1-v)$.

\subsection{Deep-RULE}
The implementation of RULE follows immediately from the bilinear form of~(\ref{eq:bilin}). Note that $\phi(y)$ is a {\it fixed\/} mapping from the space of attributes to the space of class labels. For example, if there are $Q$ binary attributes and $C$ class labels, $\phi(y)$ is a $Q$ dimensional vector that encodes the presence/absence of the $Q$ attributes in class $y$
\begin{equation}
\label{eq:semcode}
\phi_{k}(y) = \left\{
\begin{array}{cl}
1 & \mbox{if class $y$ contains attribute $k$}, \\
-1 & \mbox{if class $y$ lacks attribute $k$.} 
\end{array}
\right.
\end{equation}
We denote $\phi(y)$ the {\it semantic code of class $y$\/}. To implement~\eqref{eq:bilin} in a CNN, it suffices to use one of the popular models to compute $\theta(\x; \Theta)$, add a fully-connected layer of $Q$ units and parameters $\T$, so that $a(\x) = \T^T \theta(\x;\Theta)$ is a vector of attribute scores, and define the CNN class outputs
\begin{equation}
h(\x;\T,\Theta) = \Phi^T a(\x) = \Phi^T \T^T \theta(\x;\Theta),
\label{eq:h}
\end{equation}
where $\Phi=[\phi(1), \ldots, \phi(C)] \in \mathbb{R}^{Q\times C}$. Given a training set $\Dcal = \lbrace(\x^{(i)}, y^{(i)})_{i=1}^N\rbrace$, where $y^{(i)}$ is the class label of image $\x^{(i)}$, $\T$ and $\Theta$ are learned by minimizing 
\begin{equation}
\textstyle\Rcal[h,\Dcal] = \sum_i L\left(h(\x^{(i)};\T,\Theta), y^{(i)}\right)
\label{eq:RULErisk}
\end{equation}
where $L$ is some classification loss, typically the cross-entropy $L({\bf v},y) = -\log(\rho_y({\bf v}))$ of softmax outputs $\rho({\bf v})$.
\begin{table}[t]
\centering
\caption{Notation.}
\label{tab:notation}
\resizebox{0.99\linewidth}{!}{
	\begin{tabular}{rl}
		\toprule
		Symbol & Meaning \\ \toprule
		$\Phi$/$\Phi_{ZS}$ & Semantic codeword matrix for training/ZS classes \\
		$\phi(y)$ & Semantic codeword of class $y$ (column of $\Phi$) \\
		$\phi_k(y)$ & Semantic-state codewords (``building blocks'' of $\phi(y)$) \\ 
		$\W$ & Classification codeword matrix (related to $\Phi$ through \eqref{eq:cwdreg}) \\ 
		$\w_y$ & Classification codewords (columns of $\W$) \\
		$\Acal^\prime$ & Effective attribute space \\
		$\Acal^\prime_T$ / $\Acal^\prime_{ZS}$ & Subspace of $\Acal^\prime$ spanned by the columns of $\Phi$ / $\Phi_{ZS}$ \\
		\bottomrule 
	\end{tabular}
}
\vspace{-7pt}
\end{table}

\subsection{Relationships}
\label{sec:relations}
Both Deep-RIS and Deep-RULE have advantages and disadvantages, which can be observed by comparing the risks of \eqref{eq:RTISrisk} and \eqref{eq:RULErisk}. Since the attributes $a_k(\x)$ are the quantities of interest for ZSL, it is useful to understand how the two methods provide supervision to the space $\Acal$ of attributes. From~(\ref{eq:RTISrisk}), Deep-RIS provides supervision to the individual attributes $a_k(\x)$. Since $a_k(\cdot) = \1_k^T a(\cdot)$, where $\1_k$ is the $k^{th}$ vector in the canonical basis (1 in the $k^{th}$ position and 0 elsewhere), the supervision is along the {\it canonical directions of $\Acal$\/}. On the other hand, (\ref{eq:h})-(\ref{eq:RULErisk}) only depend on the projections $\phi(y)^Ta(\x)$ of $a(\cdot)$ along the vector encodings $\phi(\cdot)$ of all training classes. Hence, RULE only provides supervision to the {\it the column space ${\cal C}(\Phi)$ of $\Phi$.\/}

\begin{figure}[t!]
	\centering
	\begin{minipage}{\linewidth}
		\centering
		\includegraphics[width=0.5\linewidth]{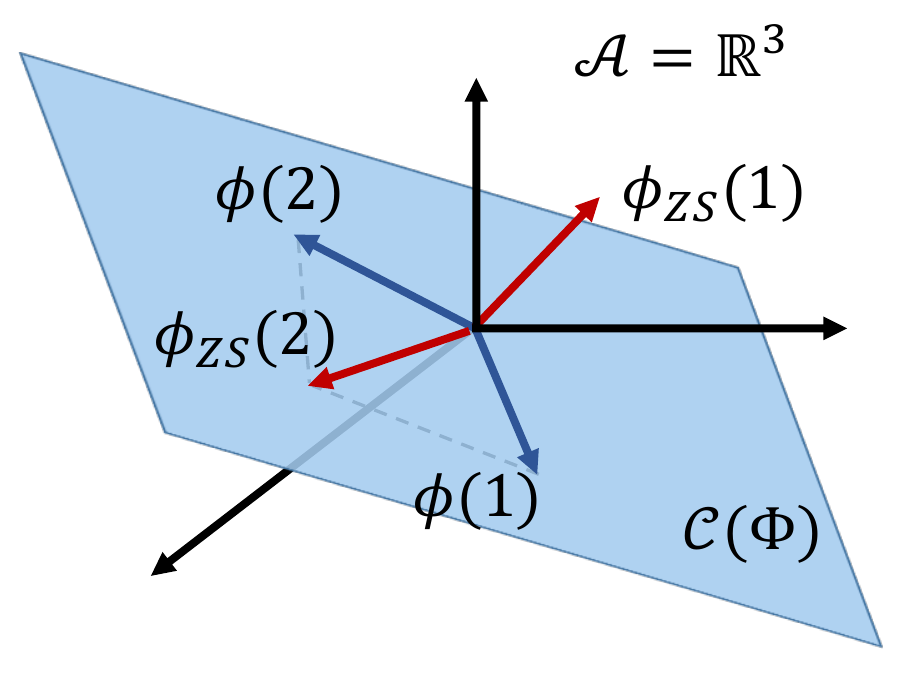}
	\end{minipage}
	\caption{Attribute space ($Q=3$). Semantic codes for two training classes shown in blue, and two ZS classes shown in red.}
	\label{fig:AttSpace}
	\vspace{-7pt}
\end{figure}

In practice, we are often on the regime of Figure~\ref{fig:AttSpace}, where the number of attributes $Q$ is larger than the number of training classes $C$. It follows that ${\cal C}(\Phi)$ can be fairly low dimensional (dimension $C$) and the left null space ${\cal N}(\Phi^T)$ fairly high dimensional (dimension $Q-C$). Hence, while RIS constraints all attributes, RULE leaves $Q-C$ attribute dimensions unconstrained. In this case, ZS classes with semantic codes $\phi_{ZS}$ misaligned with ${\cal C}(\Phi)$ cannot be expected to be accurately predicted. In the limit, RULE is completely inadequate to discriminate ZS classes when $\phi_{ZS}$ is perpendicular to ${\cal C}(\Phi)$, such as $\phi_{ZS}(1)$ in Figure \ref{fig:AttSpace}. This suggests the superiority of RIS over RULE. However, because RIS supervises attributes independently, it has no ability to learn attribute dependencies, e.g. that the attributes ``has wings'' and ``lives in the water'' have a strong negative correlation. These dependencies can be thought of as constraints that reduce the effective dimensionality of the attribute space. They imply that the attribute vectors $a(\x)$ of natural images do not span $\Acal$, but only an {\it effective attribute subspace\/} $\Acal^\prime$ of dimension $Q^\prime < Q$. By learning only on ${\cal C}(\Phi) \subset \Acal^\prime$, Deep-RULE provides supervision explicitly in this space. This suggests that Deep-RULE should outperform Deep-RIS.

Overall, the relative performance of the two approaches depends on the overlap between the subspaces of $\Acal^\prime$ covered by the training and ZS classes, denoted $\Acal^\prime_{T}$ and $\Acal^\prime_{ZS}$ respectively. If $\Acal^\prime_{T}$ contains all the directions $\phi_{ZS}$ that define $\Acal^\prime_{ZS}$, Deep-RULE will outperform Deep-RIS. If the ZS classes are defined by directions $\phi_{ZS}$ not contained in $\Acal^\prime_{T}$, Deep-RIS will likely outperform Deep-RULE.

\section{Semantically consistent regularization}
\label{sec:framework}
In this section, we introduce the {\it Semantically COnsistent REgularizer\/} (SCoRe) architecture.

\subsection{Attributes as regularization constraints}
In the previous section, we saw that the relative performance of Deep-RIS and Deep-RULE depends on the alignment between the subspaces of $\Acal^\prime$ that define the training and ZS classes, $\Acal^\prime_{T}$ and $\Acal^\prime_{ZS}$. In an ideal scenario, $\Acal^\prime_{T}=\Acal^\prime$ and so $\phi_{ZS}(c)\in\Acal^\prime_{T}$ for any ZS class $c$. However, this is unlikely to happen for datasets of tractable size, and the subsets $\Acal^\prime_{T}$ and  $\Acal^\prime_{ZS}$ are most likely not aligned. 
%This is the scenario that we consider in this work. 

Under this scenario, Deep-RIS and Deep-RULE compliment each other. 
%Both methods enforce attribute constraints during learning. 
While Deep-RIS enforces first-order constraints on the statistics of single attributes, Deep-RULE enforces second-order constraints, by constraining the statistics of linear attribute combinations. If the two strategies are combined, Deep-RULE can explain attribute dependencies that appear on both training and ZS classes, leaving to Deep-RIS the task of constraining the attribute distribution on the remainder of the space. It is, thus, natural to combine the two strategies.
We accomplish this by mapping them into regularization constraints.

\subsection{Recognition and regularization}
\label{sec:regularization}
An object recognizer maps an image $\x$ into a class
\begin{equation}
\label{eq:11}
y^* = \mathop{\arg\max}_{c\in\{1,\ldots,C\}} h_c(\x),
\end{equation}
where $h(\x) = (h_1(\x), \ldots, h_{C}(\x))$ is a vector of confidence  scores for the assignment of $\x$ to each class, and $y^*$ the class prediction. The score function  $h(\cdot)$ is usually learned by minimizing an empirical risk ${\cal R}_E[h]$, under a complexity constraint $\Omega[h]$ to improve generalization, i.e. 
\begin{equation}
h^* = \arg\min_h {\cal R}_E[h] +\lambda \Omega[h]
\label{eq:lagrangian}
\end{equation}
where $\lambda \geq 0$ is a Lagrange multiplier, and $\Omega[\cdot]$ a regularizer that favors score functions of low complexity. Common usages of $\Omega[\cdot]$ include shrinkage~\cite{Gruber1998}, sparse representations~\cite{Cheng2015} or weight decay~\cite{krogh1991}. Since all these approaches simply favor solutions of low complexity, they are a form of {\it task-insensitive\/} regularization. For ZSL, this type of regularization has indeed been used to control the variance of 1) semantic scores or 2) backward projections of object embeddings into the feature space \cite{Romera2015}, as well as to suppress noisy semantics \cite{Qiao2016}.

In this work, rather than a \textit{generic} penalty on the complexity of $h(.)$, we propose a \textit{task-sensitive} form of regularization, which favors score functions $h(\cdot)$ with the \textit{added functionality} of attribute prediction. This regularization is implemented with two complimentary mechanisms, introduced in the next two sections.

\subsection{Codeword regularization}
The first mechanism exploits the fact that the score functions of \eqref{eq:11} are always of the form
\begin{equation}
\label{eq:hc}
h_c(\x) = \left\langle \w_c,f(\x)\right\rangle,
\end{equation}
where $\langle\cdot,\cdot\rangle$ denotes an inner product, $f(\cdot)$ a {\it predictor}, and $\lbrace\w_1, \ldots, \w_C\rbrace$ a set of $C$ class {\it codewords}. We denote $\w_c$ as the {\it classification code\/} of class $c$.
For example, in binary classification, algorithms such as boosting~\cite{Freund1995} or SVM~\cite{Cortes1995} simply choose $1/-1$ as the codewords of the positive/negative class.%, and \eqref{eq:11} simplifies into {$y^* = sgn[f(\x)]$}. 
Similarly, for C-ary classification, neural networks \cite{Lecun1995} or multi-class SVMs \cite{Weston1998} rely on one-hot encodings that lead to the typical decision rule {$\textstyle y^* = \mathop{\arg\max}_{j\in\{1,\ldots,C\}} f_j(\x)$}. There is, however, no reason to be limited by these classical sets. %Different sets simply reflect different class \textit{encodings}.

By comparing the score functions of \eqref{eq:h} and \eqref{eq:hc}, Deep-RULE can be interpreted as learning the optimal predictor $a(\x)$ for a classification code given by~(\ref{eq:semcode}), i.e. $\w_c = \phi(c)$. Hence, Deep-RULE can be seen as a form of very strict CNN regularization, where the final fully-connected layer is set to these semantic codes. In general, fixing network weights is undesirable, as better results are usually obtained by learning them from data.
We avoid this by using the semantic codes $\phi(c)$ as loose regularization constraints, under the framework of \eqref{eq:lagrangian}. Similarly to Deep-RULE, we learn the predictor $f$ using cross-entropy as the empirical risk ${\Rcal}_E$, and score functions of the form
\begin{equation}
h(\x;\W,\T, \Theta) = \W^T f(\x) = \W^T \T^T \theta(\x; \Theta)
\label{eq:h2}
\end{equation}
where the columns of $\W$ contain the weight vectors $\w_c$ of the last CNN layer. This is complemented by a {\it codeword regularizer\/} 
\begin{equation}
\textstyle\Omega[\W] =\frac{1}{2} \sum_{c=1}^{C} ||\w_c - \phi(c)||^2
\label{eq:cwdreg}
\end{equation}
that favors classification codes $\w_c$ similar to the semantic codes $\phi(c)$. Note that, up to terms that do not depend on $\w_c$, this can be written as $\Omega[\W] \sim \sum_{c=1}^{C} \frac{1}{2}||\w_c||^2  - \sum_{c=1}^{C} \w_c^T\phi(c).$ In the Lagrangian of~(\ref{eq:lagrangian}), the first summation becomes the ``weight decay'' regularizer already implemented by most CNN learning packages. Thus, effectively, 
\begin{equation}
\textstyle\Omega[\W] = - \sum_{c=1}^{C} \w_c^T\phi(c).
\end{equation}
In sum, the use of codeword regularization forces the CNN to model attribute dependencies by aligning the learned classification codes $\w_c$ with semantic codes $\phi(c)$.

\subsection{Loss-based regularization}
The second mechanism, denoted \textit{loss-based regularization}, aims to constraint attributes beyond $\Acal^\prime_T$, and provides explicit regularization to attribute predictions. It is implemented by introducing an \textit{auxiliary risk} ${\cal R}_A[f]$ in the optimization, i.e. replacing ${\Rcal_E}[h]$ in \eqref{eq:lagrangian} by ${\cal R}_E[h] +\lambda {\cal R}_A[f]$ where ${\cal R}_A[f]$ is the sum of attribute prediction risks of \eqref{eq:RTISrisk}. This drives the score function to produce accurate attribute predictions, in addition to classification.

\subsection{SCoRe}
Given a training set of images $\x^{(i)}$, attribute labels $(s_1^{(i)}, \ldots,s_Q^{(i)})$, and class labels $y^{(i)}$, the regularizers of the previous sections are combined into the SCoRe objective
\begin{eqnarray}
\underset{\Theta, \T, \W}{\text{minimize}} &&\textstyle \sum_i L\left(h(\x^{(i)};\W,\T, \Theta), y^{(i)}\right) \nonumber\\
&&\textstyle+ \lambda \sum_i \sum_k L_b\left(f_k(\x^{(i)}; \bt_k, \Theta), s_k^{(i)}\right)\nonumber\\
&&\textstyle+ \beta\Omega[\W], \label{eq:lagrangianTD}
\end{eqnarray}
where $h(\cdot)$ is given by \eqref{eq:h2}, $f_k(\x; \bt_k, \Theta)=\bt_k^T\theta(\x;\Theta)$ is the $k^{th}$ semantic predictor, $\Omega[\W]$ the codeword regularizer of \eqref{eq:cwdreg}, and $\lambda$ and $\beta$ Lagrange multipliers that control the tightness of the regularization constraints. 

Depending on the value of these multipliers, SCoRe can learn a standard CNN, Deep-RIS, or Deep-RULE. When $\lambda = \beta = 0$, all the regularization constraints are disregarded and the classifier is a standard recognizer for the training classes. Increasing $\lambda$ and $\beta$ improves its transfer ability. On one hand, regardless of $\beta$, increasing $\lambda$ makes SCoRe more like Deep-RIS. In the limit of $\lambda \to \infty$, the first summation plays no role in the optimization, $\Omega$ is trivially minimized by $\w_c=\phi(c)$, and~(\ref{eq:lagrangianTD}) is reduced to the Deep-RIS optimization problem of~\eqref{eq:RTISrisk}. 
On the other hand, maintaining $\lambda = 0$ while increasing $\beta$ makes SCoRe similar Deep-RULE. For large values of $\beta$, the learning algorithm emphasizes the similarity between classification and semantic codes, trading off classification performance for semantic alignment. Finally, when both $\lambda$ and $\beta$ are non-zero, SCoRe learns the classifier that best satisfies the corresponding trade-off between the three goals: recognition, attribute predictions, and alignment with the semantic code.

\section{Semantics}
\label{sec:semantics}
In this section, we discuss the encoding of different semantics under the SCoRe framework.

\begin{figure}[t!]
	\centering
	\begin{minipage}{\linewidth}
		\centering
		\includegraphics[width=\linewidth]{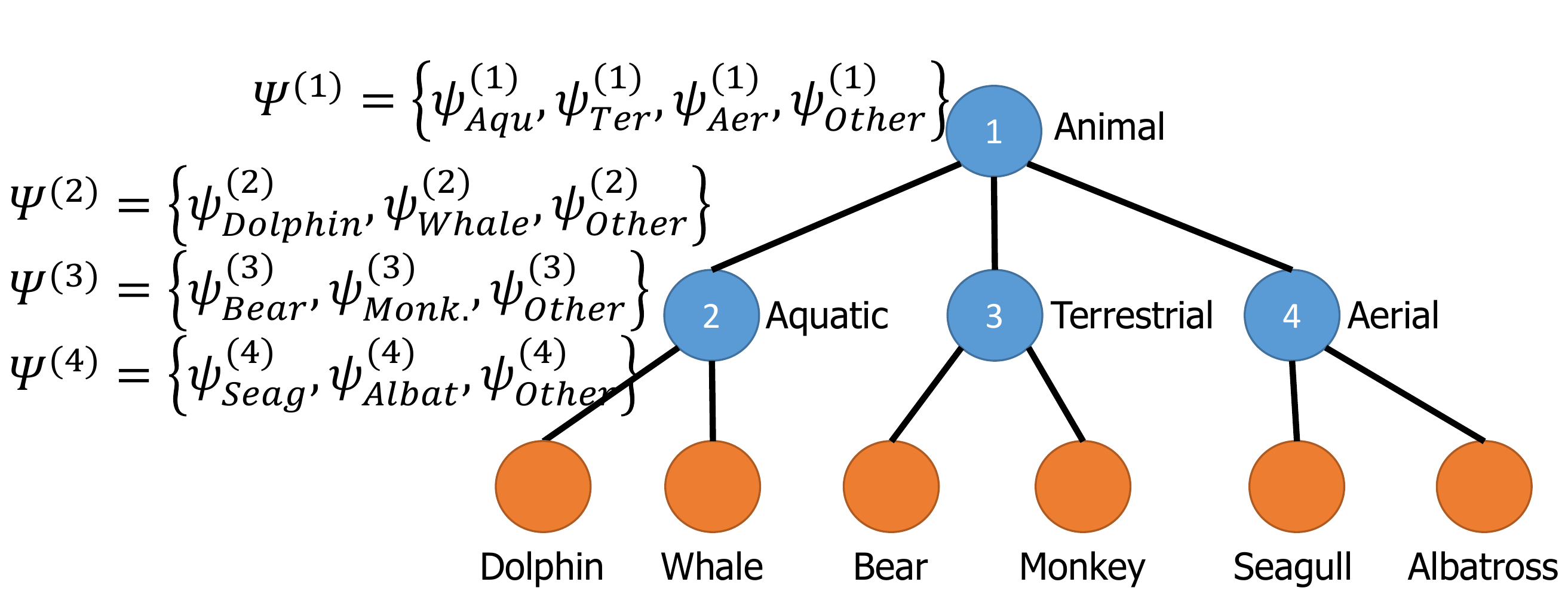}
	\end{minipage}
\vspace{3pt}
	\caption{Semantic encoding for a taxonomy of six animal classes.}
	\label{fig:hrchy}
	\vspace{-5pt}
\end{figure}

\begin{figure*}[t!]
	\centering
	\includegraphics[width=0.7\linewidth]{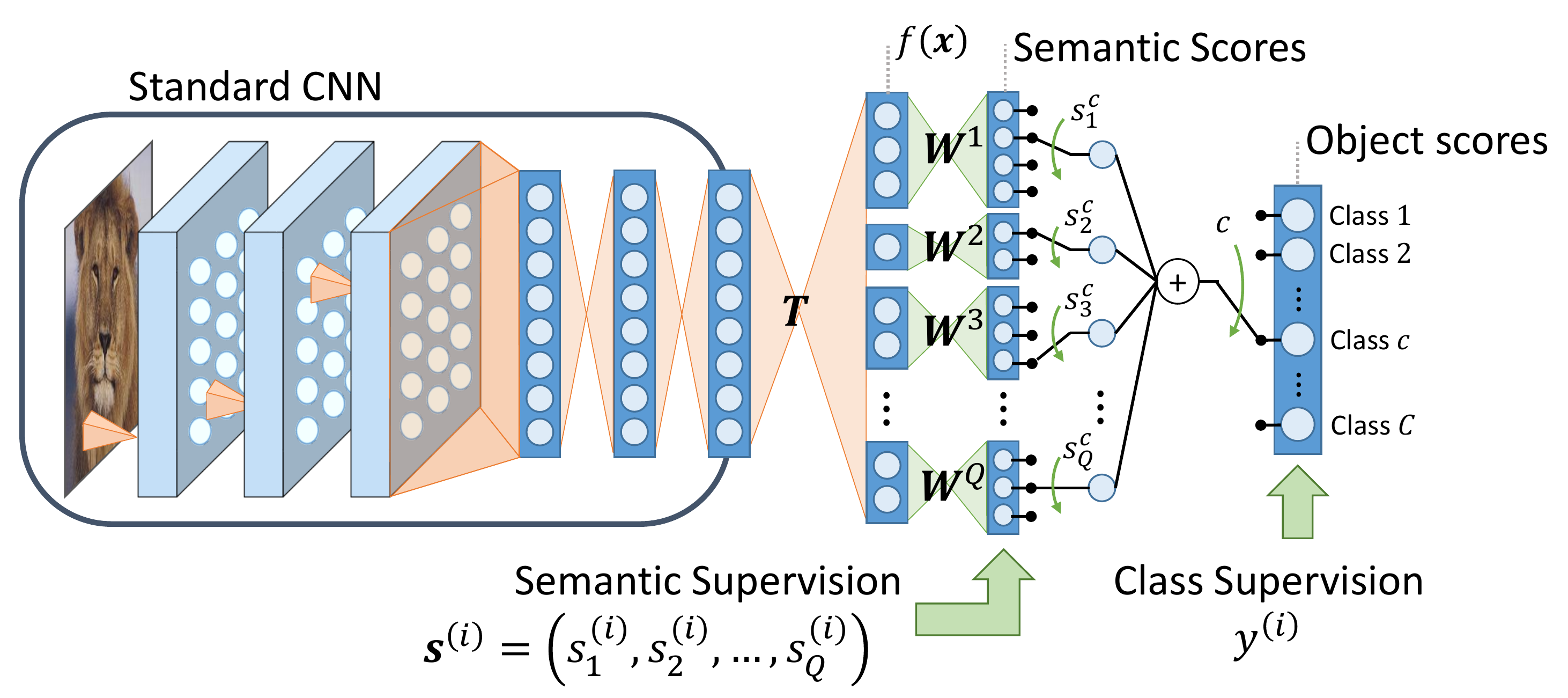}
	\caption{Deep-SCoRe. Feature extraction based on common CNN 
		architectures. Classification is performed by first computing 
		semantic scores through codewords $\W^k$, and then combining them 
		into class scores using known class/semantics relations $s^c_k$.}
	\label{fig:deepscore}
	\vspace{-7pt}
\end{figure*}

\subsection{Attributes}
So far, we assumed that semantics are binary attributes. Each attribute is mapped into an entry of the semantic code according to \eqref{eq:semcode}, which is used to represent each class, i.e.
\begin{equation}
\phi(y) = concat(\phi_1(y), \ldots, \phi_Q(y)).
\label{eq:concat1}
\end{equation}
To support different degrees of certainty on class/attribute associations, continuous attributes are also easily implemented by making $\phi_k(y) \in [-1,1]$.

\subsection{Beyond binary semantics}
\label{sec:scoremodel}
SCoRe can be easily extended to semantics with more than two states. Consider a semantic $k$ with $S_k$ states. In this case, each state is itself represented by a codeword, i.e.
\begin{equation}
\phi_k(y) \in \Psi^{(k)} = \{\psi^{(k)}_1, \ldots, \psi^{(k)}_{S_k}\},
\end{equation}
where $\psi^{(k)}_i$ are {\it semantic state codewords\/}. Then, the semantic code $\phi(y)$ of class $y$ is built by concatenating $\phi_k(y)$ for all $k$, as in \eqref{eq:concat1}. 
Similarly to the binary case, a predictor $f(\x)$ learned under this codeword set will attempt to approximate $\phi_k(y)$ for images $\x$ of class $y$. The state of the $k^{th}$ semantic can thus be recovered from $f$ with $s_k^* = \mathop{\arg\max}_{i=1, \ldots, S_k}\langle \psi_i^{(k)},f_k(\x)\rangle$ where $\psi_i^{(k)}$ is the codeword of state $i$ of the $k^{th}$ semantic, and $f_k(\cdot)$ the corresponding subspace of $f(\cdot)$. Many semantic state codewords can be defined. We now provide some examples.

\vspace{-10pt}
\paragraph{Taxonomies}
In this work, we consider taxonomic encodings that emphasize node specific decisions, by interpreting each node as a semantic concept. 
As illustrated in Figure~\ref{fig:hrchy}, a semantic state codeword set ${\Psi^{(k)}}$ is defined per node $k$. Its state codewords identify all possible children nodes plus a reject option. For example, the codeword set $\Psi^{(2)}$ of node $2$ contains codewords $\psi^{(2)}_{dolphin}$ and $\psi^{(2)}_{whale}$, plus the reject codeword $\psi^{(2)}_{other}$. Under this taxonomic encoding, the semantic code $\phi(y)$ identifies the relevance of each node to the class $y$. An internal node that is an ancestor of $y$ contributes with the codeword corresponding to the branch selection (needed to reach the class) at that node. A node that is not an ancestor contributes with the reject codeword. For example, in Figure \ref{fig:hrchy}, the class ``bear'' receives the code ${\phi(\textit{bear}) = concat\left(\psi^{(1)}_{Ter}, \psi^{(2)}_{Other}, \psi^{(3)}_{Bear}, \psi^{(4)}_{Other}\right)}$.

It remains to define the codeword sets $\Vcal^{(k)}$. These could be used to reflect further semantic information. 
In the tree of Figure \ref{fig:hrchy}, $\Vcal^{(1)}$ could encode a set of attributes that distinguish aquatic, terrestrial, and aerial animals, such as ``has fins,'' ``has legs'' or ``has wings''.
In this work, since no semantic information is available beyond the taxonomy itself, we rely on the maximally separated codeword sets of \cite{Saberian2011}. Under this procedure, a $Q$-ways decision is mapped into the set of codewords defined as the vertices of a $Q$-sided regular polygon in $Q-1$ dimensions centered at the origin.

\vspace{-10pt}
\paragraph{Word2Vec}
Word2Vec is a procedure to generate word embeddings. A word $w$ is mapped into a high-dimensional vector $\xi(w) \in \chi$ by a neural network trained from large text corpora to reconstruct linguistic contexts of words. 
For semantic annotation, this mapping is used as the semantic code, i.e.~each class $y$ is encoded by the vector $\phi(y) = \xi(y)$.% in a set $\Vcal = \{\xi(1), \ldots, \xi(C)\}$ of $C$ Word2Vec codewords. 

In this work, we use the skip-gram architecture proposed by Mikolov \etal \cite{Mikolov2013}. Its embeddings are determined by two parameters: size of the encoding layer and the window size that defines a context for each word. Rather than relying on a single model, we learn $Q$ Word2Vec embeddings $\xi_k(y), k \in \{1, \ldots, Q\}$, using $Q$ different combinations of the two parameters. This creates $Q$ codeword sets $\Vcal^{(k)}$. The semantic code then represents class $c$ by a string of the resulting vectors $\phi_k(y) =\xi_k(y)$, using~\eqref{eq:concat1}.

\section{Deep-SCoRe}
\label{sec:deep-score}
Deep-SCoRe implements \eqref{eq:h2} using a CNN to compute $\theta(\x;\Theta)$. Parameters $\Theta$, $\W$ and $\T$ are learned from \eqref{eq:lagrangianTD}, using a semantic code that combines various semantic state codeword sets $\Vcal^{(k)}$. These can be relative to attributes, taxonomy nodes, Word2Vec mappings, or any other semantic encoding. From \eqref{eq:hc}, class scores decompose into
\begin{equation}
\textstyle h_c(\x) = \sum_k h_c^{(k)}(\x) = 
\sum_k \langle \w_{s_k^c}^{(k)}, f_k(\x)\rangle
\label{eq:hc2}
\end{equation}
where $s_k^c$ is the state of the $k^{th}$ semantic under class $c$, $\w_{s_k^c}^{(k)}$ the corresponding codeword, and $f_k(\cdot)$ the corresponding subspace of $f(\cdot)$. Semantic predictions are obtained by computing the dot-products
\begin{equation}
\textstyle u_i^{(k)}(\x) =\langle\w_i^{(k)}, f_k(\x)\rangle
\label{eq:ui}
\end{equation}
for all states $i$ of semantic $k$ and choosing the state
\begin{equation}
s^*_k = \arg\max_i \textstyle u_i^{(k)}(\x).
\end{equation}
While \eqref{eq:hc2} and \eqref{eq:ui} could be computed separately, the structure of (\ref{eq:hc2}) allows shared computation. This can be accomplished by adding two layers to the semantic predictor $f(\x)$, which we denote \textit{semantic encoding} (SE) layers.

As shown in Figure \ref{fig:deepscore}, a CNN is used to compute the predictor ${f(\x)=\left(f_1,\ldots,f_Q\right)(\x)}$. Similarly to Deep-RIS and Deep-RULE, this is implemented through a linear transformation $\T$ of a feature vector $\theta(\x)$ computed with one of the popular CNN models. 
The first SE layer then consists of $Q$ parallel fully-connected layers that compute the semantic scores $u^{(k)}_i(\x)$ for each of the $Q$ semantics. The weights of each branch $k$ contain the classification codewords $\w_i^{(k)}$ and are learned under the codeword regularizer of \eqref{eq:cwdreg}. 
The second SE layer then selects, for each class $c$, a single output from each branch $k$ corresponding to the state $s_k^c$ of the $k^{th}$ semantic of class $c$. These outputs are added to obtain the class recognition score $h_c(\x)$.
This is easily implemented by a fully connected layer of predetermined sparse weights of 0s and 1s that remain fixed throughout training.

\vspace{-10pt}
\paragraph{Learning:} 
Consider a training set of three-tuples: (a) the image $\x^{(i)}$; (b) the vector of semantic states ${\s^{(i)}}$; and (c) the class label ${y^{(i)}}$. As shown in Figure~\ref{fig:deepscore}, the state vectors ${\s^{(i)}}$ are used as supervisory signals for the first SE layer and the labels ${y^{(i)}}$ as supervisory signals for the second. These supervisory signals and the semantic codes $\phi(y)$ are used to compute the Lagrangian risk of \eqref{eq:lagrangianTD}, and all parameters are optimized by back-propagation using Caffe toolbox \cite{jia2014}.

Deep-SCoRe models were trained by fine tuning pre-trained CNNs using stochastic gradient descent (SGD) with momentum of 0.9 and weight decay of 0.0005. The learning rate was chosen empirically for each experiment.

\section{Experiments}
\label{sec:experiments}
In this section, we discuss several experiments carried out to evaluate the ZSL performance of Deep-SCoRe. Source code is available at \url{https://github.com/pedro-morgado/score-zeroshot}.

\subsection{Experimental setup}

\paragraph{Datasets:}
Three datasets were considered: Animals with Attributes \cite{Lampert2009} (AwA), Caltech-UCSD Birds 200-2011 \cite{Wah2011} (CUB), and a subset of the Imaging FlowCytobot \cite{Sosik2007} (IFCB) dataset. Table \ref{tab:datasets} summarizes their statistics. On AwA and CUB, the partition into source and target classes for ZSL is as specified by \cite{Lampert2009} and \cite{Akata2013}, respectively. On IFCB, which is now first used for ZSL, classes were partitioned randomly. A separate set of validation classes (10/50/6 for the AwA/CUB/IFCB datasets, respectively) was also drawn randomly to tune SCoRe parameters. 

\vspace{-10pt}
\paragraph{Image representation:}
Images were resized to $256\times256$ pixels, with the exception of IFCB, where aspect ratios differ widely and resizing introduces considerable distortion. Instead, each image was first resized along the longest axis and the shortest axis then padded with the average pixel value, to preserve the aspect ratio. Typical data augmentation techniques were used for training (random cropping and mirroring), and the center crop was used for testing. 
Three CNN architectures were used to implement $\theta(\x)$: AlexNet \cite{Krizhevsky2012} (layer fc7), GoogLeNet \cite{Szegedy2014} (layer pool5) and VGG19 \cite{Simonyan2014} (layer fc7).

\vspace{-10pt}
\paragraph{Semantics:}
Three sources of semantics were evaluated.

\textit{Visual attributes:} Continuous attributes have been shown to be superior to their binary counterparts and were used on AwA and CUB. On IFCB, where no attributes were defined previously, a list of 35 visual attributes was assembled and annotated by an expert with binary labels, using several sources from the oceanographic community \cite{WoRMS2015,Cassis2015}.

\textit{Taxonomies} were created by pruning the WordNet tree~\cite{Miller1995} for the training and ZS classes, and eliminating dummy nodes containing a single child. In the rare situations where WordNet was not fine-grained enough to distinguish between a set of classes, the taxonomy was expanded by simply assigning each object into its own leaf. 

\textit{Word2Vec} models were trained on a Wikipedia archive, dated June 1st, 2016. Three different window sizes ($3$, $5$ and $10$) and vector dimensions ($50$, $100$ and $500$) were used, leading to a total of 9 Word2Vec codeword sets.

\begin{table}[t]
	\centering
	\caption{Summary of dataset statistics.}
	\label{tab:datasets}
	\resizebox{0.9\linewidth}{!}{
		\begin{tabular}{rcccc}
			\toprule
			Dataset & Images & \begin{tabular}{c}Train/ZS\\
				Classes\end{tabular} & Attributes & \begin{tabular}{c}Hierarchy\\Source\end{tabular} \\ \toprule
			AwA     & 30,475 & 40/10  & 85  & WordNet \cite{Miller1995} \\ \midrule
			CUB     & 11,788 & 150/50 & 312 & WordNet \cite{Miller1995} \\ \midrule
			IFCB    & 28,853 & 22/8   & 35  & --- \\ \bottomrule 
		\end{tabular}
	}
	\vspace{-5pt}
\end{table}

\begin{figure}[t!]
	\begin{picture}(0,190)
	\put(7,95){\includegraphics[width=0.9\linewidth]{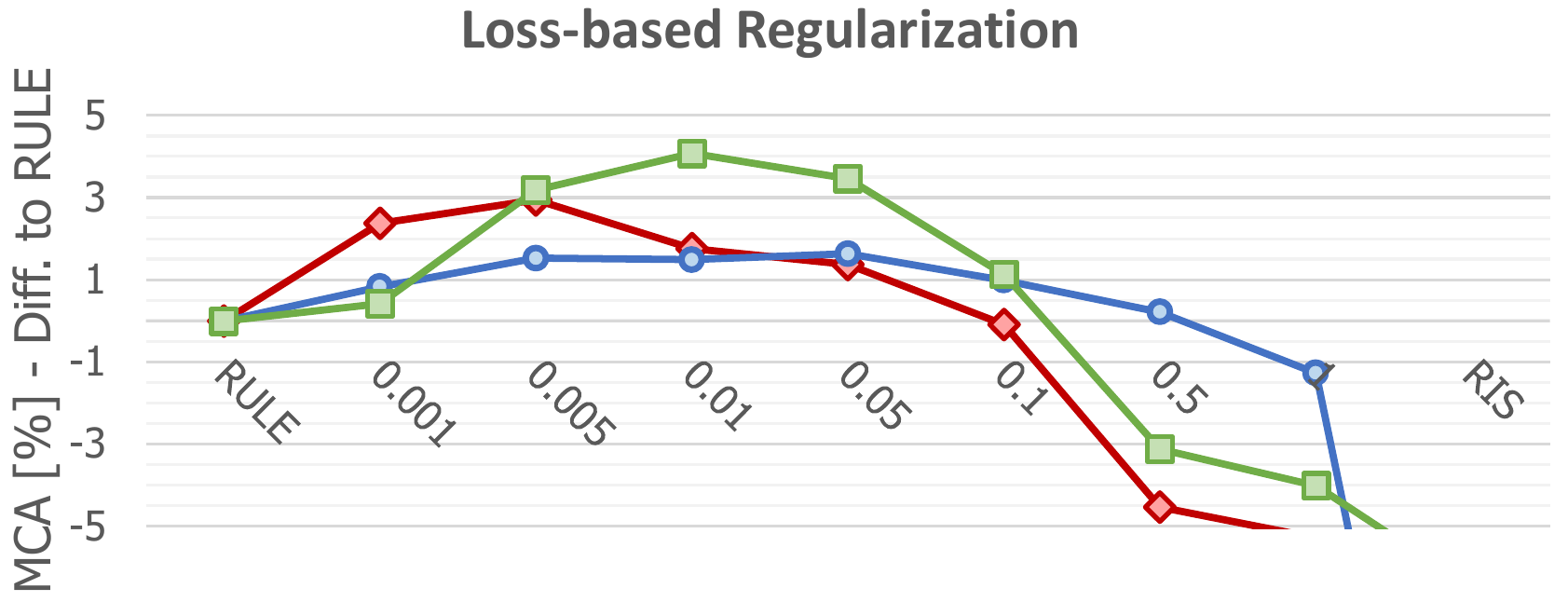}}
	\put(10,0){\includegraphics[width=0.9\linewidth]{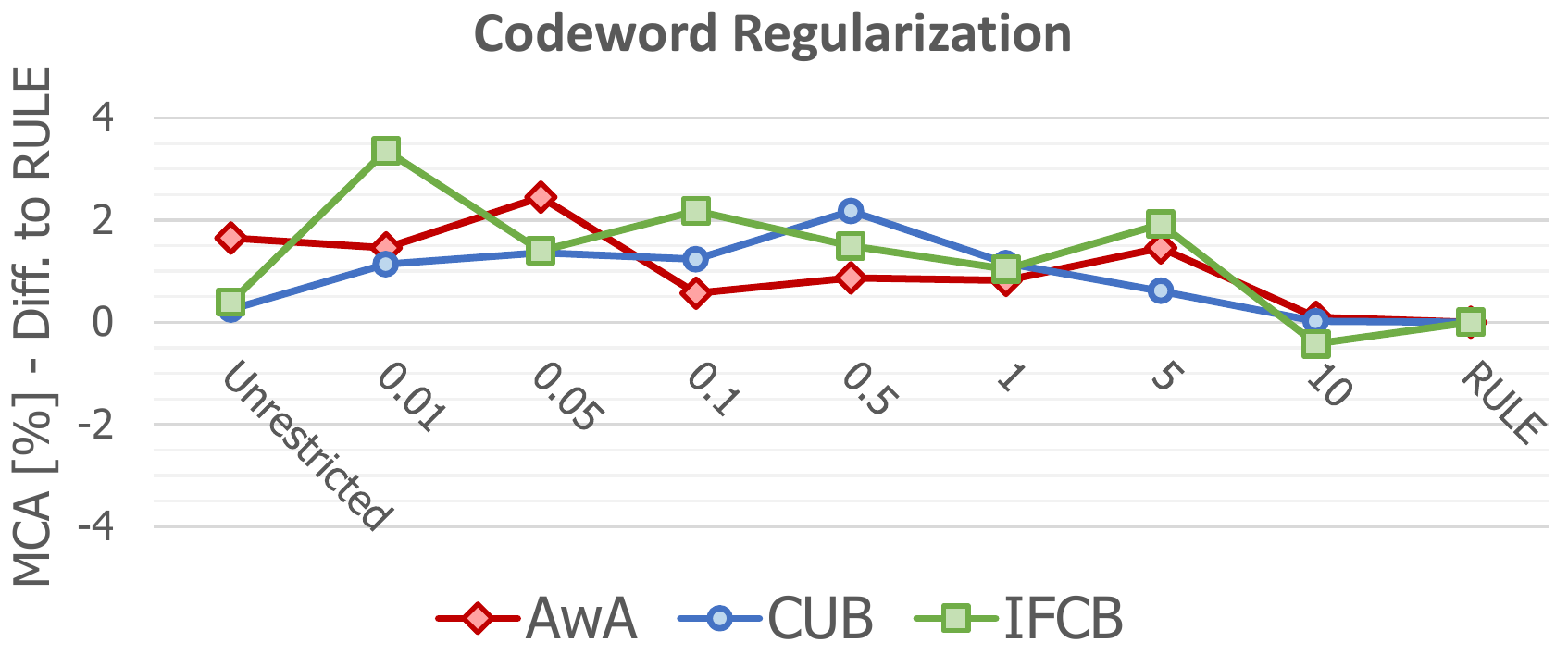}}
	\put(110,20){$\beta$}\put(110,107){$\lambda$}
	\end{picture}
	\caption{Influence of loss-based and codeword regularization on ZSL. Absolute improvement over RULE in ZS-MCA.}
	\label{fig:regul}
	\vspace{-5pt}
\end{figure}

\subsection{Results}

\paragraph{Gains of regularization:}
We started by evaluating codeword and loss-based regularization. The importance of the two regularizers was assessed separately on all datasets using visual attributes and GoogLeNet. In both cases, we measured the gains over Deep-RULE, in which classification codewords are set to $\w_c=\phi(c)$ and $\lambda=0$. The gains of loss-based regularization were evaluated by increasing $\lambda$ while keeping $\beta = 0$. Under this setting, the classifier converges to Deep-RIS in the limit of $\lambda \rightarrow \infty$. Conversely, the gains of codeword regularization were measured by increasing $\beta$ while keeping $\lambda = 0$. In this case, the classifier converges to an unrestricted object recognizer when $\beta = 0$ and to Deep-RULE when $\beta \rightarrow \infty$. Figure \ref{fig:regul} presents the absolute improvement in ZS mean class accuracy (ZS-MCA) over Deep-RULE, as a function of the Lagrange multipliers.

\begin{table}[t!]
	\centering
	\caption{ZS-MCA[\%] of various methods. 
		\mbox{A - AlexNet \cite{Krizhevsky2012};} 
		\mbox{G - GoogLeNet \cite{Szegedy2014};} 
		\mbox{V - VGG19 \cite{Simonyan2014}.}}
	\label{tab:StateArt}
	\resizebox{0.48\textwidth}{!}{
		\begin{tabular}{r|ccc|ccc}
			\midrule
			& \multicolumn{3}{c}{AwA}
			& \multicolumn{3}{|c}{CUB} \\ 
			\midrule
			& A & G & V & A & G & V \\
			\midrule
			DAP \cite{Lampert2013} & 45.3${}^\dagger$ & 59.5${}^\ddagger$ &-
			& 16.9${}^\dagger$ & 36.6${}^\ddagger$ & - \\
			SJE \cite{Akata2015} & 61.9 & 66.7 &- & 40.3&50.1& - \\
			ES-ZSL${}^\S$ \cite{Romera2015} &53.0&74.2&74.4 & 40.6&53.1&49.0 \\
			Huang \etal\cite{Huang2015} & 45.6 &-&-& 17.5 &-&- \\
			Liang \etal\cite{Liang2015} & 48.6 &-&-& 18.2 &-&- \\
			Changpinyo \etal\cite{Changpinyo2016} &-& 72.9&-&-& 54.7 &- \\
			Xian \etal\cite{Xian2016}&- & 72.5&-&- & 45.6 &- \\
			Zhang \etal\cite{Zhang2015} &-&-& 76.3 &-&-& 30.4 \\
			Gan \etal\cite{Gan2016} &-&-& 73.8 &-&-& 43.7 \\\midrule
			Deep-RIS   &    56.6 &    68.9 &    66.4 &    24.3 &    37.5 &    39.1 \\
			Deep-RULE  &    65.3 &    76.3 &    78.0 &    46.0 &    57.1 &    57.9 \\
			Deep-SCoRe &\bf 66.7 &\bf 78.3 &\bf 82.8 &\bf 48.5 &\bf 58.4 &\bf 59.5 \\
			\bottomrule
			\multicolumn{7}{l}{${}^\dagger$As reported by Liang \etal\cite{Liang2015}.
				${}^\ddagger$As reported by Al-Halah \etal\cite{AlHalah2016}.}\\
			\multicolumn{7}{l}{${}^\S$Self implementation.}
	\end{tabular}}
	\vspace{-7pt}
\end{table}

Both regularizers produced gains over Deep-RULE with absolute gains as high as $3$ ZS-MCA points. This demonstrates the importance of learning the classification codewords, rather than fixing them. Note that, for codeword regularization, best results were obtained for intermediate values of $\beta$, which encourage consistency between the semantic and classification codes, but leave enough flexibility to learn a classification code superior to its semantic counterpart.
In all cases, the MCA of SCoRe was much superior to that of RIS, confirming the importance of modeling attribute dependencies through the first term of~\eqref{eq:lagrangianTD}. Finally, SCoRe performance was also superior to that of the unrestricted CNN. This demonstrates the benefits of regularization. Interestingly, this was {\it not\/} the case of RIS, which always underperformed the unrestricted CNN, or RULE that only achieved on par results in CUB and IFCB\footnote{The unrestricted CNN is initialized with semantic codes. If random initialization was used, ZSL would not be possible.}.

In Section~\ref{sec:relations}, we hypothesized that loss-based regularization becomes more important as the alignment between the subspaces of $\Acal^\prime$ spanned by training and ZS classes decreases. To test this hypothesis, we measured this alignment by computing the average orthogonal distance between the semantic codeword $\phi(c)$ of each ZS class and the subspace spanned by the codewords of training classes.
The average distances were $0.1244$ for CUB, $0.3063$ for AwA, and $0.4181$ for IFCB, indicating that the transfer is easiest for CUB and hardest for IFC. This is consistent with the plots of Figure~\ref{fig:regul}, which show largest gains of loss-based regularization on IFCB followed by AwA and then CUB.

\vspace{-10pt}
\paragraph{Comparisons to state-of-the-art methods:}
A comparison to the literature is not trivial since methods differ in 1) CNN implementation, 2) train/ZS class partitioning, and 3) semantic space representation. 
To mitigate these differences, we focused on attribute semantics which have most available results. Methods that use alternative semantics~\cite{Qiao2016,Fu2016,Reed2016,Akata2016b} or that use unlabeled images from ZS classes for training \cite{Fu2014a,Fu2015,Li2015b,Kodirov2015} were disregarded for this comparison. Deep-SCoRe hyper-parameters $\lambda$ and $\beta$ were tuned on a subset of the training classes.

Table \ref{tab:StateArt} compares our ZS-MCA to previous approaches using three CNN architectures: AlexNet, GoogLeNet and VGG19. Although results vary drastically with CNN, it is clear that Deep-SCoRe outperforms all previous approaches on all datasets, achieving impressive gains over the state-of-the-art for every architecture: 4.8\%, 4.1\% and 6.5\% on AwA and 7.9\%, 3.7\% and 10.5\% on CUB with AlexNet, GoogLeNet and VGG19, respectively. 

\begin{figure}[t!]
	\begin{picture}(0,135)(0,0)
	\put(20,5){\includegraphics[height=125pt]{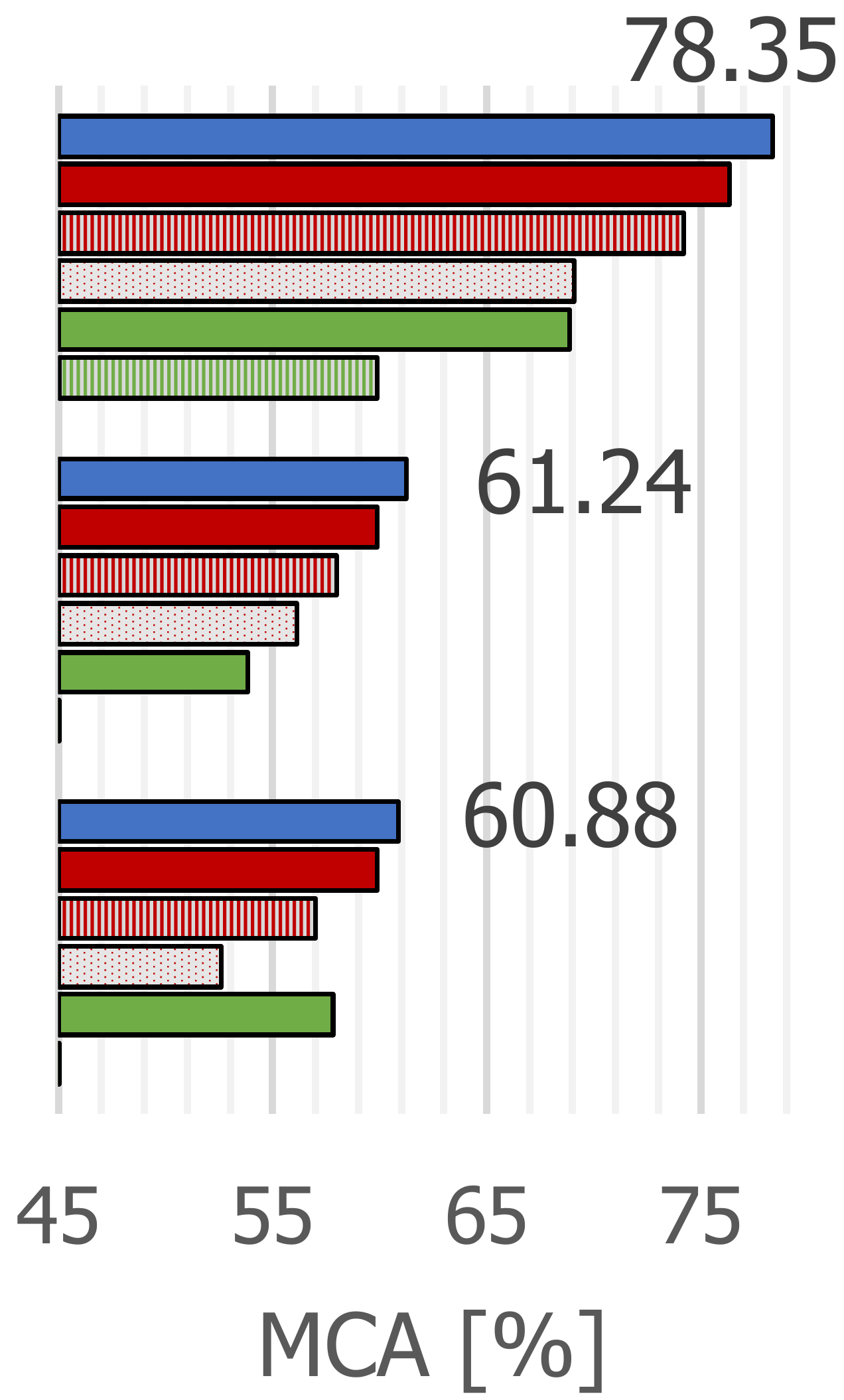}}
	\put(110,5){\includegraphics[height=125pt]{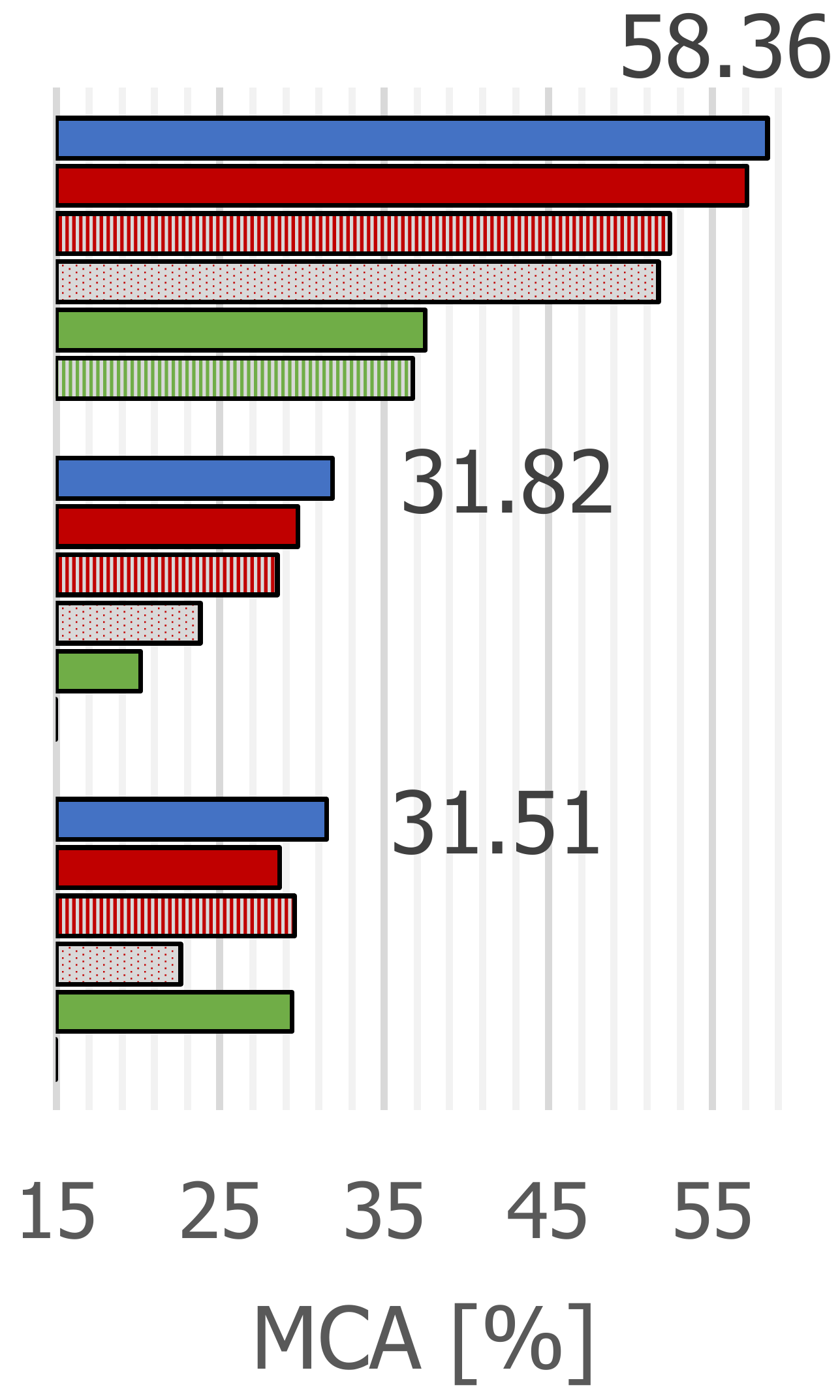}}
	%\put(155,30){\includegraphics[height=110pt]{figs/semIFCB}}
	\put(190,20){\includegraphics[height=100pt]{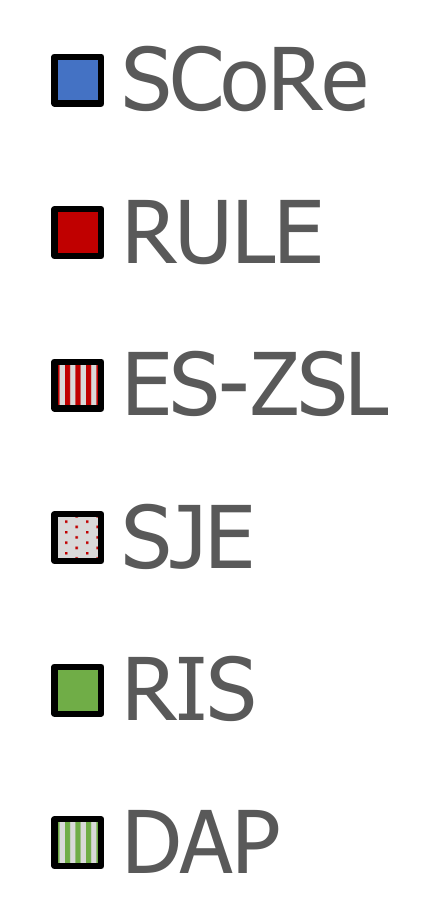}}
	\put(10,45){\small$\mathcal{W}$}
	\put(10,75){\small$\mathcal{H}$}
	\put(10,105){\small$\mathcal{A}$}
	\put(50,130){\bf AwA}
	\put(140,130){\bf CUB}
	%\put(180,142){\bf IFCB}
	\end{picture}
	\caption{ZSL performance using different semantics. $\mathcal{A}$ -- Attributes; $\mathcal{H}$ -- Hierarchies; $\mathcal{W}$ -- Word2Vec. DAP results reported in \cite{AlHalah2015}. SJE and ES-ZSL self-implemented.}
	\label{fig:semantics}
	\vspace{-7pt}
\end{figure}
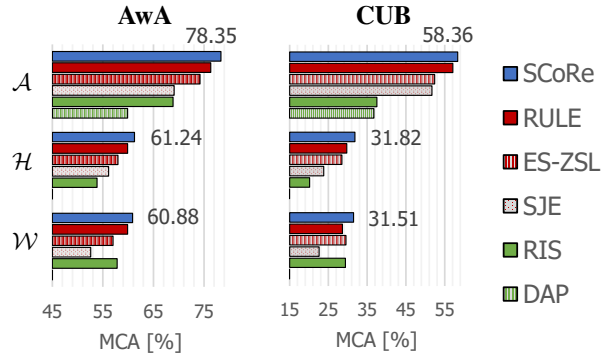

\vspace{-10pt}
\paragraph{Multiple semantics:}
We finally studied the performance of Deep-SCoRe with attributes, taxonomies, and Word2Vec embeddings. Figure~\ref{fig:semantics} compares Deep-SCoRe and its variants to popular RIS and RULE approaches in the literature: DAP~\cite{Lampert2009} (RIS), SJE \cite{Akata2015} and ES-ZSL \cite{Romera2015} (RULE). All approaches were implemented with the semantic codes of Section \ref{sec:semantics}. The best results, which were all obtained with Deep-SCoRe, are also shown. Figure \ref{fig:semantics} supports two main conclusions.
First, as shown in~\cite{Akata2016,Xian2016,Changpinyo2016}, attributes enable by far the most effective transfer. This is not surprising since attributes tend to be discriminant properties of the various object classes. Taxonomies or Word2Vec are most informative of grouping or contextual information.
Second, while all approaches rely on regularization, the nature of this regularization matters. The task-sensitive regularization of Deep-SCoRe always outperformed the task-insensitive regularization of ES-ZSL, and the combination of loss-based and codeword regularization (Deep-SCoRe) always outperformed a fixed semantic code (Deep-RULE and SJE) or loss-based regularization (Deep-RIS and DAP).

\section{Conclusion}\vspace{-3pt}
In this work, we analyzed the type of supervision provided by previous approaches. The complementarity found between class and semantic supervision lead to the introduction of a new ZSL procedure, denoted SCoRe, where a CNN is learned together with a semantic codeword set and two forms of semantic constraints: loss-based and codeword regularization. State-of-the-art zero-shot performance was achieved in various datasets.

{\small

\vfill\pagebreak

\bibliographystyle{ieee}
\bibliography{refs_final}

\begin{thebibliography}{10}\itemsep=-1pt

\bibitem{Akata2016b}
Z.~Akata, M.~Malinowski, M.~Fritz, and B.~Schiele.
\newblock Multi-cue zero-shot learning with strong supervision.
\newblock In {\em Computer Vision and Pattern Recognition (CVPR), IEEE
  Conf.~on}, 2016.

\bibitem{Akata2013}
Z.~Akata, F.~Perronnin, Z.~Harchaoui, and C.~Schmid.
\newblock {Label-embedding for attribute-based classification}.
\newblock In {\em Computer Vision and Pattern Recognition (CVPR), IEEE
  Conf.~on}, 2013.

\bibitem{Akata2016}
Z.~Akata, F.~Perronnin, Z.~Harchaoui, and C.~Schmid.
\newblock Label-embedding for image classification.
\newblock {\em Pattern Analysis and Machine Intelligence (TPAMI), IEEE
  Trans.~on}, 38(7):1425--1438, 2016.

\bibitem{Akata2015}
Z.~Akata, S.~Reed, D.~Walter, H.~Lee, and B.~Schiele.
\newblock Evaluation of output embeddings for fine-grained image
  classification.
\newblock In {\em Computer Vision and Pattern Recognition (CVPR), IEEE
  Conf.~on}, 2015.

\bibitem{AlHalah2015}
Z.~Al-Halah and R.~Stiefelhagen.
\newblock {How to transfer? Zero-shot object recognition via hierarchical
  transfer of semantic attributes}.
\newblock In {\em Applications of Computer Vision, IEEE Winter Conf.~on}, 2015.

\bibitem{AlHalah2016}
Z.~Al-Halah, M.~Tapaswi, and R.~Stiefelhagen.
\newblock Recovering the missing link: Predicting class-attribute associations
  for unsupervised zero-shot learning.
\newblock In {\em Computer Vision and Pattern Recognition (CVPR), IEEE
  Conf.~on}, 2016.

\bibitem{Cassis2015}
D.~Cassis.
\newblock Phytopedia - the phytoplankton encyclpaedia project.
\newblock Available at: \url{http://www.eos.ubc.ca/research/phytoplankton/}.
\newblock Accessed: 2015-11-04.

\bibitem{Changpinyo2016}
S.~Changpinyo, W.-L. Chao, B.~Gong, and F.~Sha.
\newblock Synthesized classifiers for zero-shot learning.
\newblock In {\em Computer Vision and Pattern Recognition (CVPR), IEEE
  Conf.~on}, 2016.

\bibitem{Chen2014}
C.-Y. Chen and K.~Grauman.
\newblock Inferring analogous attributes.
\newblock In {\em Computer Vision and Pattern Recognition (CVPR), IEEE
  Conf.~on}, 2014.

\bibitem{Chen2012}
H.~Chen, A.~Gallagher, and B.~Girod.
\newblock Describing clothing by semantic attributes.
\newblock In {\em Computer Vision (ECCV), European Conf.~on}, 2012.

\bibitem{Cheng2015}
H.~Cheng.
\newblock {\em Sparse representation, modeling and learning in visual
  recognition}.
\newblock Springer, 2015.

\bibitem{Cortes1995}
C.~Cortes and V.~Vapnik.
\newblock Support-vector networks.
\newblock {\em Machine learning}, 20(3):273--297, 1995.

\bibitem{Deng2009}
J.~Deng, W.~Dong, R.~Socher, L.-J. Li, K.~Li, and L.~Fei-Fei.
\newblock {ImageNet: A large-scale hierarchical image database}.
\newblock In {\em Computer Vision and Pattern Recognition (CVPR), IEEE
  Conf.~on}, 2009.

\bibitem{Farhadi2009}
A.~Farhadi, I.~Endres, D.~Hoiem, and D.~Forsyth.
\newblock {Describing objects by their attributes}.
\newblock In {\em Computer Vision and Pattern Recognition (CVPR), IEEE
  Conf.~on}, 2009.

\bibitem{Freund1995}
Y.~Freund and R.~E. Schapire.
\newblock A desicion-theoretic generalization of on-line learning and an
  application to boosting.
\newblock In {\em Computational Learning Theory, European Conf.~on}. Springer,
  1995.

\bibitem{Frome2013}
A.~Frome, G.~S. Corrado, J.~Shlens, S.~Bengio, J.~Dean, T.~Mikolov, et~al.
\newblock Devise: A deep visual-semantic embedding model.
\newblock In {\em Advances in Neural Information Processing Systems (NIPS)},
  2013.

\bibitem{Fu2014a}
Y.~Fu, T.~M. Hospedales, T.~Xiang, and S.~Gong.
\newblock {Transductive multi-view zero-shot recognition and annotation}.
\newblock In {\em Computer Vision (ECCV), European Conf.~on}, 2014.

\bibitem{Fu2015}
Y.~Fu, T.~M. Hospedales, T.~Xiang, and S.~Gong.
\newblock Transductive multi-view zero-shot learning.
\newblock {\em Pattern Analysis and Machine Intelligence (TPAMI), IEEE
  Trans.~on}, 37(11):2332--2345, 2015.

\bibitem{Fu2016}
Y.~Fu and L.~Sigal.
\newblock Semi-supervised vocabulary-informed learning.
\newblock In {\em Computer Vision and Pattern Recognition (CVPR), IEEE
  Conf.~on}, 2016.

\bibitem{Fu2015b}
Z.~Fu, T.~Xiang, E.~Kodirov, and S.~Gong.
\newblock Zero-shot object recognition by semantic manifold distance.
\newblock In {\em Computer Vision and Pattern Recognition (CVPR), IEEE
  Conf.~on}, 2015.

\bibitem{Gan2016}
C.~Gan, T.~Yang, and B.~Gong.
\newblock Learning attributes equals multi-source domain generalization.
\newblock In {\em Computer Vision and Pattern Recognition (CVPR), IEEE
  Conf.~on}, 2016.

\bibitem{Gruber1998}
M.~Gruber.
\newblock {\em Improving Efficiency by Shrinkage: The James--Stein and Ridge
  Regression Estimators}, volume 156.
\newblock CRC Press, 1998.

\bibitem{Huang2015}
S.~Huang, M.~Elhoseiny, A.~Elgammal, and D.~Yang.
\newblock {Learning hypergraph-regularized attribute predictors}.
\newblock {\em arXiv}, 2015.

\bibitem{hwang2011}
S.~J. Hwang, F.~Sha, and K.~Grauman.
\newblock Sharing features between objects and their attributes.
\newblock In {\em Computer Vision and Pattern Recognition (CVPR), IEEE
  Conf.~on}, pages 1761--1768, 2011.

\bibitem{Jayaraman2014}
D.~Jayaraman and K.~Grauman.
\newblock {Zero-shot recognition with unreliable attributes}.
\newblock In {\em Advances in Neural Information Processing Systems (NIPS)},
  2014.

\bibitem{jia2014}
Y.~Jia, E.~Shelhamer, J.~Donahue, S.~Karayev, J.~Long, R.~Girshick,
  S.~Guadarrama, and T.~Darrell.
\newblock Caffe: Convolutional architecture for fast feature embedding.
\newblock {\em arXiv:1408.5093}, 2014.

\bibitem{Kodirov2015}
E.~Kodirov, T.~Xiang, Z.~Fu, and S.~Gong.
\newblock Unsupervised domain adaptation for zero-shot learning.
\newblock In {\em Computer Vision and Pattern Recognition (CVPR), IEEE
  Conf.~on}, 2015.

\bibitem{Krizhevsky2012}
A.~Krizhevsky, I.~Sutskever, and G.~E. Hinton.
\newblock Imagenet classification with deep convolutional neural networks.
\newblock In {\em Advances in Neural Information Processing Systems (NIPS)},
  2012.

\bibitem{krogh1991}
A.~Krogh and J.~A. Hertz.
\newblock A simple weight decay can improve generalization.
\newblock In {\em Advances in Neural Information Processing Systems (NIPS)},
  1991.

\bibitem{Kumar2009}
N.~Kumar, A.~C. Berg, P.~N. Belhumeur, and S.~K. Nayar.
\newblock {Attribute and simile classifiers for face verification}.
\newblock In {\em Computer Vision (ICCV), IEEE International Conf.~on}, 2009.

\bibitem{Lampert2009}
C.~H. Lampert, H.~Nickisch, and S.~Harmeling.
\newblock {Learning to detect unseen object classes by between-class attribute
  transfer}.
\newblock In {\em Computer Vision and Pattern Recognition (CVPR), IEEE
  Conf.~on}, 2009.

\bibitem{Lampert2013}
C.~H. Lampert, H.~Nickisch, and S.~Harmeling.
\newblock {Attribute-based classification for zero-shot visual object
  categorization}.
\newblock {\em Pattern Analysis and Machine Intelligence (TPAMI), IEEE
  Trans.~on}, 36(3), 2013.

\bibitem{Lecun1995}
Y.~LeCun and Y.~Bengio.
\newblock Convolutional networks for images, speech, and time series.
\newblock {\em The handbook of brain theory and neural networks}, 3361(10),
  1995.

\bibitem{Li2010}
L.-J. Li, H.~Su, L.~Fei-Fei, and E.~P. Xing.
\newblock Object bank: A high-level image representation for scene
  classification \& semantic feature sparsification.
\newblock In {\em Advances in Neural Information Processing Systems (NIPS)},
  2010.

\bibitem{Li2015}
X.~Li and Y.~Guo.
\newblock Max-margin zero-shot learning for multi-class classification.
\newblock In {\em Artificial Intelligence and Statistics (ICAIS), International
  Conf.~on}, 2015.

\bibitem{Li2015b}
X.~Li, Y.~Guo, and D.~Schuurmans.
\newblock Semi-supervised zero-shot classification with label representation
  learning.
\newblock In {\em Computer Vision (ICCV), IEEE International Conf.~on}, 2015.

\bibitem{Liang2015}
K.~Liang, H.~Chang, S.~Shan, and X.~Chen.
\newblock A unified multiplicative framework for attribute learning.
\newblock In {\em Computer Vision (ICCV), IEEE International Conf.~on}, 2015.

\bibitem{WoRMS2015}
J.~Mees, G.~Boxshall, M.~Costello, et~al.
\newblock {World Register of Marine Species (WoRMS)}.
\newblock Available at: \url{http://www.marinespecies.org}.
\newblock Accessed: Dec-2016.

\bibitem{Mikolov2013}
T.~Mikolov, I.~Sutskever, K.~Chen, G.~S. Corrado, and J.~Dean.
\newblock Distributed representations of words and phrases and their
  compositionality.
\newblock In {\em Advances in Neural Information Processing Systems (NIPS)},
  2013.

\bibitem{Miller1995}
G.~A. Miller.
\newblock {WordNet: A lexical database for English}.
\newblock {\em Communications of the ACM}, 38(11):39--41, 1995.

\bibitem{Norouzi2013}
M.~Norouzi, T.~Mikolov, S.~Bengio, Y.~Singer, J.~Shlens, A.~Frome, G.~S.
  Corrado, and J.~Dean.
\newblock Zero-shot learning by convex combination of semantic embeddings.
\newblock {\em arXiv:1312.5650}, 2013.

\bibitem{Parikh2011}
D.~Parikh and K.~Grauman.
\newblock {Relative attributes}.
\newblock In {\em Computer Vision (ICCV), IEEE International Conf.~on}, 2011.

\bibitem{Qiao2016}
R.~Qiao, L.~Liu, C.~Shen, and A.~v.~d. Hengel.
\newblock Less is more: zero-shot learning from online textual documents with
  noise suppression.
\newblock In {\em Computer Vision and Pattern Recognition (CVPR), IEEE
  Conf.~on}, 2016.

\bibitem{Rasiwasia2007}
N.~Rasiwasia, P.~J. Moreno, and N.~Vasconcelos.
\newblock {Bridging the gap: Query by semantic example}.
\newblock {\em Multimedia, IEEE Trans.~on}, 9(5):923--938, 2007.

\bibitem{Rasiwasia2012}
N.~Rasiwasia and N.~Vasconcelos.
\newblock Holistic context models for visual recognition.
\newblock {\em Pattern Analysis and Machine Intelligence (TPAMI), IEEE
  Trans.~on}, 34(5):902--917, 2012.

\bibitem{Rastegari2012}
M.~Rastegari, A.~Farhadi, and D.~Forsyth.
\newblock {Attribute discovery via predictable discriminative binary codes}.
\newblock In {\em Computer Vision (ECCV), European Conf.~on}, 2012.

\bibitem{Reed2016}
S.~Reed, Z.~Akata, B.~Schiele, and H.~Lee.
\newblock Learning deep representations of fine-grained visual descriptions.
\newblock In {\em Computer Vision and Pattern Recognition (CVPR), IEEE
  Conf.~on}, 2016.

\bibitem{Rohrbach2011}
M.~Rohrbach, M.~Stark, and B.~Schiele.
\newblock {Evaluating knowledge transfer and zero-shot learning in a
  large-scale setting}.
\newblock In {\em Computer Vision and Pattern Recognition (CVPR), IEEE
  Conf.~on}, 2011.

\bibitem{Romera2015}
B.~Romera-Paredes and P.~Torr.
\newblock An embarrassingly simple approach to zero-shot learning.
\newblock In {\em Machine Learning (ICCV), International Conf.~ on}, pages
  2152--2161, 2015.

\bibitem{Saberian2011}
M.~J. Saberian and N.~Vasconcelos.
\newblock {Multiclass boosting: Theory and algorithms}.
\newblock In {\em Advances in Neural Information Processing Systems (NIPS)},
  2011.

\bibitem{Simonyan2014}
K.~Simonyan and A.~Zisserman.
\newblock Very deep convolutional networks for large-scale image recognition.
\newblock {\em arXiv preprint arXiv:1409.1556}, 2014.

\bibitem{Sosik2007}
H.~M. Sosik and R.~J. Olson.
\newblock Automated taxonomic classification of phytoplankton sampled with
  imaging-in-flow cytometry.
\newblock {\em Limnology and Oceanography: Methods}, 5(6):204--216, 2007.

\bibitem{Su2010}
Y.~Su, M.~Allan, and F.~Jurie.
\newblock {Improving object classification using semantic attributes}.
\newblock In {\em British Machine Vision Conference (BMVC)}, 2010.

\bibitem{Szegedy2014}
C.~Szegedy, W.~Liu, Y.~Jia, P.~Sermanet, S.~Reed, D.~Anguelov, D.~Erhan,
  V.~Vanhoucke, and A.~Rabinovich.
\newblock Going deeper with convolutions.
\newblock {\em arXiv preprint arXiv:1409.4842}, 2014.

\bibitem{Torresani2010}
L.~Torresani, M.~Szummer, and A.~Fitzgibbon.
\newblock Efficient object category recognition using classemes.
\newblock In {\em Computer Vision (ECCV), European Conf.~on}, 2010.

\bibitem{Vogel2007}
J.~Vogel and B.~Schiele.
\newblock Semantic modeling of natural scenes for content-based image
  retrieval.
\newblock {\em Computer Vision, International Journal of}, 72(2):133--157,
  2007.

\bibitem{Wah2011}
C.~Wah, S.~Branson, P.~Welinder, P.~Perona, and S.~Belongie.
\newblock {The Caltech-UCSD Birds-200-2011 Dataset}.
\newblock Technical Report CNS-TR-2011-001, California Institute of Technology,
  2011.

\bibitem{Wang2013}
X.~Wang and Q.~Ji.
\newblock {A unified probabilistic approach modeling relationships between
  attributes and objects}.
\newblock In {\em Computer Vision (ICCV), IEEE International Conf.~on}, 2013.

\bibitem{Weston1998}
J.~Weston and C.~Watkins.
\newblock Multi-class support vector machines.
\newblock Technical report, Citeseer, 1998.

\bibitem{Xian2016}
Y.~Xian, Z.~Akata, G.~Sharma, Q.~Nguyen, M.~Hein, and B.~Schiele.
\newblock Latent embeddings for zero-shot classification.
\newblock In {\em Computer Vision and Pattern Recognition (CVPR), IEEE
  Conf.~on}, 2016.

\bibitem{Zhang2015}
Z.~Zhang and V.~Saligrama.
\newblock Zero-shot learning via semantic similarity embedding.
\newblock In {\em Computer Vision and Pattern Recognition (CVPR), IEEE
  Conf.~on}, 2015.

\end{thebibliography}
}
\end{document}